\documentclass{article}
\usepackage{amsmath}
\usepackage{amssymb}
\usepackage{amsthm}
\usepackage{algorithm}
\usepackage{algpseudocode}
\usepackage{appendix}
\usepackage{authblk}
\usepackage[margin=1.4in]{geometry}
\usepackage{hyperref} 
\usepackage{graphicx}
\usepackage{url}
\newtheorem{definition}{Definition}
\title{From Logic to Language: A Trust Index for Problem Solving with LLMs}
\author{
    Tehseen Rug\thanks{Corresponding author: tehseen.rug@iteratec.com}, Felix Böhmer, Tessa Pfattheicher \\
    iteratec GmbH \\
    St.-Martin-Str. 114 \\
    81669 Munich, Germany
}
\date{}

\begin{document}

\maketitle

\begin{abstract}
Classical computation, grounded in formal, logical systems, has been the engine of technological progress for decades, excelling at problems that can be described with unambiguous rules. This paradigm, however, leaves a vast ocean of human problems---those characterized by ambiguity, dynamic environments, and subjective context---largely untouched. The advent of Large Language Models (LLMs) represents a fundamental shift, enabling computational systems to engage with this previously inaccessible domain using natural language. This paper introduces a unified framework to understand and contrast these problem-solving paradigms. We define and delineate the problem spaces addressable by formal languages versus natural language. 
While solutions to the former problem class can be evaluated using binary quality measures, the latter requires a much more nuanced definition of approximate solution space taking into account the vagueness, subjectivity and ambiguity inherent to natural language.  
We therefore introduce a vector-valued trust index $Q$, which reflects solution quality and distinguishes the binary correctness of formal solutions from the continuous adequacy spectrum characteristic of natural language solutions. Within this framework, we propose two statistical quality dimensions. Normalized bi-semantic entropy measures robustness and conceptual diversity of LLM answers given semantic variation in problem formulations. Emotional valence maps subjective valuation of a solution to a quantifiable metric that can be maximized by invoking statistical measures.
The concepts introduced in this work will provide a more rigorous understanding of the capabilities, limitations, and inherent nature of problem-solving in the age of LLMs.
\end{abstract}

\section{Introduction}
The Church-Turing thesis posits that any function computable by an algorithm can be computed by a Turing machine. This principle underpins the digital world, defining the realm of the computationally solvable. Programs operating on well-defined rules produce verifiably correct outputs, which is the strength of the classical paradigm.

However, many real-world problems do not come with a known or fixed algorithmic specification. Therefore, the set of human-relevant problems far exceeds the set of problems that are computable in the classical sense. Consider navigating an unprecedented financial crisis, boosting team morale, or negotiating a delicate diplomatic exchange. These scenarios lack predefined rules or sufficient historical data. Instead of turning to formal algorithms, we thus rely on communication, planning, and contextual understanding—primarily through natural language. The goal then is not to find the one perfect solution that typically is out of reach in such cases, but to figure out solutions that are "good enough" in the sense that they lead to outcomes appropriate for our needs. 

Natural language problem-solving offers extraordinary flexibility: it enables us to address virtually any situation expressible in words. Yet this power comes at a cost. Natural language is inherently ambiguous, context-sensitive, and subjective. Solutions are not simply “right” or “wrong” but judged along multiple dimensions—from factual accuracy to emotional resonance and social acceptability. Defining what makes a solution "good" thus becomes a challenging task. 
Typically, we characterize the quality of a solution by whether it aligns with social norms, reflects majority consensus, or—when viewed in retrospect—leads to positive outcomes such as increased revenue, to name just a few examples. In other words, solution quality is multifaceted and largely context-dependent. Nevertheless, in principle, there are measures that allow us to distinguish "good" from "bad" solutions depending on the problem at hand.

While natural language-based problem solving is deeply rooted in human capabilities, Large Language Models (LLMs), as the first computational systems to meaningfully enter this domain, have drastically changed this perspective. By automatically processing and generating natural language, they can address problems that are difficult or impossible to formulate with the rigid logic of code.
At the same time, the inherent challenges in assessing the quality of natural language solutions are propagated from human to digital systems, making the evaluation of LLM-based applications notoriously more challenging as compared to formal systems.

Recent literature such as \cite{zheng2023judging, ragas, geval} proposes specific evaluation metrics. Further, work such as \cite{boundedrationality} acknowledges a nuanced perspective on solution quality when evaluating LLM-based systems. Focusing on primary objectives, the authors provide insights into alignment between human-based and LLM-based decision-making offering a new perspective on effectively judging solution quality in an operational setting. Still, there is yet no comprehensive formal framework to define what constitutes problem-solving in natural language systems and what distinguishes it from formal systems. Given the capabilities enabled by generative AI, we believe that such a framework is highly beneficial for multiple reasons:

\begin{itemize}
    \item It provides a common language for researchers and developers, guiding the design of models, benchmarks and applications with shared assumptions and goals.
    
    \item For real-world deployment (e.g., law, medicine, business), we need transparent standards for what “good” solutions are. A formal framework helps define thresholds guiding the development of systems that are "good enough".
    
    \item Natural language is the medium of human cognition and collaboration. Formalizing how machines use it for problem-solving helps bring them closer to meaningful human-AI interaction. For example, an AI tutor helping students with essay writing must balance factual accuracy, tone, and pedagogical clarity. By formalizing these quality dimensions, we enable more effective and human-aligned educational interactions.
\end{itemize}
The main purpose of this paper is to bridge the gap between intuitive understanding of how problems can be addressed using natural language and a rigorous, unified framework for assessing solution quality through quantifiable and statistically accessible metrics. While this work is primarily theoretical in nature, our goal is to establish a formal framework that provides conceptual clarity and guidance for future empirical studies.
Our contributions include:
\begin{itemize}
    \item A conceptual delineation of problem spaces for formal systems versus those addressable by natural language and LLMs. We thereby provide a universal language suitable for classifying the potential of LLMs in real-world problem-solving thereby offering a complementary view to the theoretical work on computability of LLMs provided in \cite{llmsandturing}.  
    
    \item The construction of a trust index $Q$ as a measure of solution quality, contrasting the “exactness” of logical solutions with the “adequacy” of linguistic ones.
    
    \item The construction of two new statistically accessible realizations of $Q$ serving as examples for semantic and emotional aspects of solutions.
    
    \item A toy model application illustrating how to ground these concepts in real-world scenarios.
\end{itemize}

By establishing this framework, we can better analyze the unique challenges and opportunities presented by LLMs and guide the development of more sophisticated and aligned AI systems.

\section{The Landscape of Problem Solving}
\label{sec:problem_spaces}
The theory of problem-solving has deep roots in cognitive science. The work by Newell and Simon \cite{newell} pioneered the conceptualization of problem-solving as a search process through an internal task-dependent problem space. Building on these ideas---and recognizing the need to distinguish computational tasks solvable by formal systems from those addressable by LLM-based problem-solving---we begin by introducing \( U \), the universe of all conceivable problems which can be framed using language. We can identify distinct (though potentially overlapping) subsets based on how they can be addressed or solved.

\subsection{The Space of Formally Solvable Problems \( P_{\text{Formal}} \)}

A problem \( p \in P_{\text{Formal}} \) if it can be characterized by:

\begin{itemize}
    \item A well-defined input set \( I \).
    \item A well-defined output set \( O \).
    \item A set of explicit, logical rules or an algorithm \( A \) such that for any \( i \in I \), \( A(i) = o \in O \), and the correctness of \( o \) can be objectively verified.
    \item \textbf{Examples:} Sort a list in ascending order, Compile code, Execute a payment in an e-commerce platform
\end{itemize}

These are the traditional targets of computer science. Programming languages are designed to express \( A \). The size of this space is theoretically vast (all computable functions), but practically constrained by our ability to specify \( A \).

\subsection{The Space of Natural Language Addressable Problems \( P_{\text{NL}} \)}

A problem \( p \in P_{\text{NL}} \) if its primary means of definition, exploration, and/or solution involves natural language\footnote{While any solution to a formal problem can, in principle, be paraphrased in natural language, we explicitly do not include such representational restatements in the definition of \( P_{\text{NL}} \).}. These problems often exhibit:

\begin{itemize}
    \item Ill-defined or evolving goals.
    \item Implicit or incomplete rules.
    \item Dependence on context, common sense, or subjective judgment.
    \item Solutions that are ``plans,'' ``strategies,'' ``narratives,'' or ``communications'' rather than deterministic outputs.
    \item \textbf{Examples:} Devise a strategy to improve team morale, Respond to an unexpected PR crisis, Write an intriguing blog article
\end{itemize}
Historically, \( P_{\text{NL}} \) was almost exclusively the domain of human cognition, as naive attempts to model such problems using rule-based logic failed to capture their dynamic nature and the subtlety of linguistic nuance.

\subsection{The Emerging Role of LLMs: \( P_{\text{LLM}} \)}

We propose that LLMs enable computational access to a significant portion of \( P_{\text{NL}} \) and also offer a new modality for interacting with some problems in \( P_{\text{Formal}} \). While it would be intriguing to postulate \( P_{\text{LLM}} \) = \( P_{\text{NL}} \), analyses such as \cite{llmsofwhat} point out aspects inherent to human language that can not yet be captured by current-day LLM-systems\footnote{An even more radical view would be to postulate that the set of problems addressable by LLMs and the set of problems addressable by natural language share common ground but are distinct, with neither being entirely contained within the other $P_{\text{LLM}} \cap P_{\text{NL}} \neq \emptyset$, $P_{\text{LLM}} \not\subseteq P_{\text{NL}}$, $P_{\text{NL}} \not\subseteq P_{\text{LLM}}$. This implies that LLMs can solve problems that are not amenable to human-language solutions. It's important to note that while reinforcement learning with verifiable rewards indicates LLMs might solve problems without employing typical human linguistic patterns (e.g. through language-mixing) \cite{deepseek}, humans can still approach these same problems using natural or formal language. Thus, based on current understanding, today's LLMs---even those designed for reasoning---haven't yet addressed problems that are fundamentally beyond what humans can accomplish with language.}. As of now we thus take the more conservative perspective. 

\begin{itemize}
    \item \( P_{\text{LLM}} \subseteq P_{\text{NL}} \): LLMs can process, generate, and reason (to an extent) with natural language, allowing them to tackle problems previously outside computational reach.
    \item \( P_{\text{LLM}} \cap P_{\text{Formal}} \ne \emptyset \): LLMs can be used to generate code, explain formal concepts in natural language, or act as natural language interfaces to formal systems — for example, through tool invocation or function calling.
\end{itemize}
The key assertion is that LLMs significantly expand the computationally addressable problem space beyond \( P_{\text{Formal}} \) into regions of \( P_{\text{NL}} \) that were, for all practical purposes, previously inaccessible to machines.

We can visualize this as: $P_{\text{Formal}} \subset (P_{\text{Formal}} \cup P_{\text{LLM}}) \subseteq U$. The  critical expansion is: $P_{\text{NL,LLM}} = P_{\text{LLM}} \setminus P_{\text{Formal}}$, i.e. the subset of U where digitalization becomes possible for the first time by virtue of LLMs. This set offers immense flexibility for automating non-formal problem-solving, albeit at the cost of the ambiguities and subjectivity inherent to natural language.

\begin{figure}[htbp] 
    \centering 
    \includegraphics[width=0.89\textwidth]{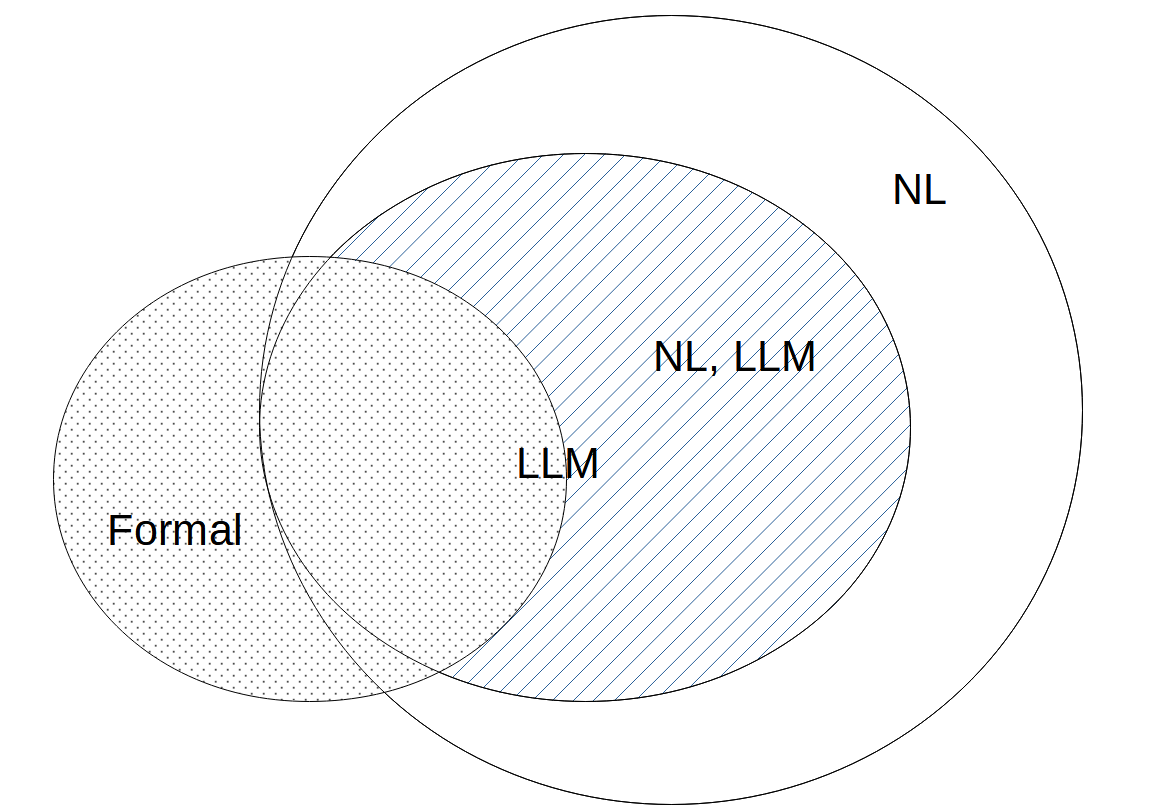} 
    \caption{Conceptual relationships between $P_{\text{Formal}}$, $P_{\text{NL}}$ and $P_{\text{LLM}}$ as defined in Section~\ref{sec:problem_spaces}.}
    \label{fig:Venn} 
\end{figure}

\vspace{1em}
In Figure~\ref{fig:Venn}, we visualize these problem spaces using a Venn diagram. It shows the delineation between problems primarily addressable via natural language or LLMs versus those solvable using formal algorithms. Section~\ref{sec:examples} will later provide concrete examples for each subregion of the full problem space \( U \). Before that, we introduce formal and quantifiable measures for evaluating the quality of any given solution.

\section{Trust Index: Formalizing Solution Quality}
\label{sec:quality}
The quality of a solution for a formal task can be judged using binary logic. In contrast, non-formal problem-solving requires a much more nuanced evaluation. To bridge this gap, we propose a universal solution quality function---referred to as trust index---denoted by $Q$, which maps any solution to a vector representing its performance across multiple quality dimensions.

\begin{definition}[Trust Index $Q$]
We define a solution quality function, $Q$, which maps a solution $s$ for a given problem $p \in U$ to a vector in the N-dimensional unit hypercube\footnote{Note that in general $Q$ also depends on provided context. We implicitly absorb such dependence by redefining $p$ accordingly so that it contains the relevant context.}.
$$Q(s,p) \rightarrow [0,1]^N$$
where $N$ is the number of quality dimensions relevant to the problem (e.g., factuality, clarity, safety, emotional tone). A value of 1 in a dimension represents hypothetical perfection in that aspect, while 0 represents complete failure\footnote{There is no objective notion of perfection for metrics such as clarity. For these types of metrics, a value of 1 should be seen as a de facto, unreachable upper limit. The practical goal, therefore, is to achieve a value that is “good enough". Below we will provide a more extensive discussion regarding what it means for a solution to be "good enough".}. 
\end{definition}
This vector-based definition allows us to elegantly formalize the fundamental difference between the problem-solving paradigms:
\begin{itemize}
    \item For formal problems $p_f \in P_{\text{Formal}}$: The quality of a solution is binary and one-dimensional; it is simply correct or incorrect\footnote{In real-world scenarios we might judge other aspects such as code quality. Using automated scanners this can be reframed as multiple formal problems (e.g. "is a block of code longer than N lines", "does the code contradict the defined styleguide" etc. can all be answered formally). Non-formalizable aspects of code quality such as an architect's "personal style" are ill-defined from a formal perspective making them part of non-formal evaluation.}. Thus, we can represent this with $N=1$, and the range of $Q$ becomes the discrete set $\{0,1\}$. The objective is to find a solution $s_f$ such that $Q(s_f, p_f)=1$.
    
    \textit{Example}: For the problem "Sort the list [3, 1, 4] in ascending order!", a program returning [1, 3, 4] has Q=1, while any other output has Q=0.

    \item For LLM-based natural language problems $p_{\text{nl,llm}} \in P_{\text{NL,LLM}}$: For these ill-defined problems, solutions exist on a spectrum of "goodness" across multiple dimensions. A perfect score of `(1, 1, ..., 1)` that satisfies all observers may be conceptually impossible. Therefore, the continuous space $[0,1]^N$ is relevant. The objective is to find a solution $s_{\text{nl,llm}}$ that is "good enough", a concept which sounds vague but which we shall formalize and operationalize later in this work.   

    \textit{Example}: For the problem "Draft an email to employees about a potential merger," we might care about three dimensions ($N=3$): clarity, reassurance, and professionalism. A clear but alarming draft might be rated $Q=(0.9, 0.2, 0.9)$. A vague but reassuring draft might be rated $Q=(0.4, 0.8, 0.7)$. There is no single solution with $Q=(1, 1, 1)$.
\end{itemize}
In the following discussion, we will focus on LLM-based natural language problems. For this set, we denote the components of the trust index $Q$ as $Q_k$ for $k=1,...,N$. To judge the quality of a solution, we must evaluate the joint distribution of this vector. The evaluation strategy for each component $Q_k$ must be tailored to its specific nature. For example, `factuality` $Q_1$ might be assessed against a knowledge base, while `reassurance` $Q_2$ requires subjective assessment. These methods must be rooted in statistical ideas for two reasons. First, LLMs themselves are statistical. Second, criteria such as readability or reassurance are subjective. We should therefore consider an ensemble of human or LLM-based evaluators, drawn from a distribution representative of the target audience. Note that the hypothesis that LLM-based evaluators can be used as proxies of a representative group of humans for the task of sentiment and emotional judgement is supported by a recent paper \cite{sentiment2025agents}. It investigates the alignment of humans and AI agents with rich profiles based on said humans in sentiment analysis tasks. The results show great alignment between responses of the humans and their AI analogues. Related work also supports the hypothesis that LLMs enriched with rich personalities can simulate human behavior, see e.g. \cite{simulacra, personalityevaluation}\footnote{Intuitively, these results are not entirely surprising. LLMs are trained on vast amounts of text data created by humans. Further, methods such as reinforcement learning from human feedback (RLHF) \cite{RHLF} guide alignment of LLM answers with human preferences. Nevertheless, checking alignment for a specific use case is still important as relevant data and personas might be sampled from different underlying distributions than data and evaluators used in pretraining and RLHF finetuning steps.}.
While these results guide the path towards using LLM-based proxies in real-world evaluation scenarios, we warn the reader not to overly rely on this methodology without justifying it given the use case under consideration. 
Such justification is feasible by evaluating alignment between users and their LLM-proxies or iterative refinement of evaluation results using human feedback (see also section \ref{sec:good enough}). 

That said, as this paper is primarily concerned with the theoretical foundations of problem and solution spaces in the context of LLMs, we do not attempt an explicit alignment study here.

\section{Concrete Examples}
\label{sec:examples}
Let us illustrate the concept of problem and solution spaces on the basis of concrete examples.  

\subsection{Example 1: The Black Swan Stock Market Event}
\textbf{Problem Type}: $P_{\text{NL}}$, partially addressable by $P_{\text{LLM}}$.
\begin{itemize}
    \item \textbf{Formal Approach}: Algorithmic trading systems based on historical data fail because the event, by definition, is outside their model. Their $Q$ drops from 1 (within model) to 0 (in crisis).
    \item \textbf{Human/LLM Approach}: Focus is on sense-making and damage control. LLMs can rapidly summarize news and sentiment to inform human decision-makers. They can draft initial communications (memos, press releases) for human review. Adjusting trading strategy accordingly might lead to a solution that is "good enough".
\end{itemize}

\subsection{Example 2: Designing a Marketing Campaign for a Novel Product}
\textbf{Problem Type}: $P_{\text{NL}}$, partially addressable by $P_{\text{LLM}}$.
\begin{itemize}
    \item \textbf{Formal Approach}: Limited to analyzing existing data or A/B testing predefined creatives. Cannot generate the core creative strategy.
    \item \textbf{Human/LLM Approach}: A highly creative and communicative process. LLMs can be used to brainstorm diverse slogans, ad copy, and marketing angles. These can, for example, be evaluated along subjective trust index dimensions such as memorability, persuasiveness, and originality.
\end{itemize}

\subsection{Example 3: Code Generation}
\textbf{Problem Type}: Problems that lie in the intersection \( P_{\text{Formal}} \cap P_{\text{NL}} \)
\begin{itemize}
    \item \textbf{Formal Approach}: Based on a programmer's precise understanding of the task, they write executable code that solves the problem. The outcome can be evaluated using binary logic—e.g., does the code sort a list correctly? If yes, $Q = 1$; otherwise, $Q = 0$.
    
    \item \textbf{Human/LLM Approach}: A user describes functionality in natural language (e.g., "Sort this list of numbers"). The LLM interprets this prompt and generates code. Since such prompts often contain implicit assumptions or ambiguity (e.g., ascending or descending order), interpreting the user's intent becomes part of the solution process. Thus, the problem-solving modality relies on both natural language interpretation and formal execution, placing this problem within both \( P_{\text{Formal}} \) and \( P_{\text{NL}} \).
\end{itemize}

\subsection{Example 4: RSA Key Generation}

\textbf{Problem Type}: Problems that are addressed purely in formal terms, i.e., \( P_{\text{Formal}} \setminus P_{\text{NL}} \)

\begin{itemize}
    \item \textbf{Formal Approach}: RSA key generation involves selecting two large prime numbers and computing their product, totient, and modular inverse to produce a public-private key pair. These steps are entirely deterministic and governed by well-defined mathematical rules. The problem has clear input, a formal algorithm, and a binary correctness condition ($Q = 1$ or $Q = 0$).
    
    \item \textbf{Human/LLM Approach}: While the task can be described in natural language (e.g., "Generate an RSA key pair"), natural language plays no essential role in solving the problem. The solution does not require contextual understanding, interpretation, or subjective reasoning. It is purely computational. Therefore, despite being expressible in natural language, RSA key generation is not addressable via natural language in the sense defined in this framework, and thus does not belong to \( P_{\text{NL}} \).
\end{itemize}

\section{Formalizing "Good Enough"}
\label{sec:good enough}
So far, we have introduced the abstract concepts of problem spaces and their corresponding solutions, along with the notion of a trust index as a measure of solution quality. Ultimately, however, the key objective is to determine whether a given solution is “good enough.” 

Although this phrase may sound informal, we adopt it here as a precise, operational concept. In our context, “good enough” should not be interpreted as “mediocre,” but rather as “correct within the bounds of uncertainty and ambiguity inherent to natural language.”

For formal problems, “good enough” collapses to identifying the one correct solution—verifiable through binary logic. In contrast, for natural language problems, the analysis is inherently more complex and lacks standardized formalism. Our goal in this section is to close this gap by providing a rigorous definition of “good enough” that can serve as a guiding principle for the design and evaluation of LLM-based systems.

For each component $Q_k$ of the trust index the ultimate goal is to optimize the expected quality $E[Q_k(s_{\text{nl,llm}})]$. However, as the task at hand is grounded in statistical concepts, we often want to do so under the boundary conditions that other statistical parameters such as variance 
$V[Q_k(s_{\text{nl,llm}})]$
stay within predefined bounds. Note that depending on the setting and task, statistical variance can emerge in different ways, e.g.
\begin{itemize}
    \item Evaluator variance: When human evaluators or LLM-based judges are used, variation may arise from individual preferences. If the evaluator pool is representative of the target user base, high variance implies that some users may find the solution unsatisfactory. Minimizing such variance helps ensure broader acceptability.
    \item Generative variance: When multiple outputs are sampled from the LLM for the same task (e.g., generating a blog article), the distribution of results may vary widely. While some variation is expected—especially for creative tasks—high-quality solutions should still be consistent with regard to certain dimensions (e.g., coherence or factuality), implying low variance in those dimensions.
\end{itemize}
As the solution space is ambiguous no such concept as an objective global optimum can be properly defined. We thus need to take a pragmatic approach. We must define what is "good enough".

Now suppose you fix an LLM's weights, its system prompt and the statistical evaluation strategy when generating a solution. Then we will have $E[Q_k(s_{\text{nl,llm}})] = q_k$ and $V[Q_k(s_{\text{nl,llm}})] = v_k$. We can normalize the expectation value of the solution quality such that $0 \leq q_k \leq 1$ for all $k$.
Based on these preliminaries, we can now define what it means for a solution to be "good enough".
\begin{definition}["Good Enough"]
Let $\hat{q}_k$ be a threshold value for $q_k$ and $\hat{v}_k$ be a threshold for $v_k$. Then we define "good enough" as follows:
$$q_k >= \hat{q}_k, v_k <= \hat{v}_k$$ for all k.
\end{definition}
Given this definition, three key questions arise:
\begin{itemize}
    \item How to compute the values $q_k$ and $v_k$ in practice?
    \item Which measures to take for adjusting these values?
    \item How to set threshold values $\hat{q}_k$ and $\hat{v}_k$?
\end{itemize}
The first question was already addressed at the end of the Section~\ref{sec:quality}. Depending on $k$ we need to construct a statistical evaluation strategy based on multiple LLM-based solution generations and/or introducing ensembles of evaluators (human-based or LLM-based). Note that in order rely on this ensemble, it should mimic the actual distribution of personas which are supposed to work with the LLM-based digitalization solution. In practice, this requires deep understanding of the domain and the user base. This understanding can be gained e.g. by polls or by including feedback mechanisms into the application.

As for the second question---adjusting $q_k$ and $v_k$---this involves modifying the behavior of the underlying LLM. This can be achieved via prompt engineering or fine-tuning, followed by re-evaluation. Since human-in-the-loop evaluation is costly, we propose a two-stage approach. First, construct LLM-as-a-Judge instances tailored to the quality dimension in question. These could include persona-driven LLMs assessing emotional impact or simulated editors judging textual appropriateness, to give some examples. Optimization is performed with respect to these proxy evaluators before final validation with human judges.

While this is a practical strategy, the final question left unanswered is: "When is it good enough?", i.e. how to set threshold values $\hat{q}_k$ and $\hat{v}_k$. Again, we take a pragmatic approach. Aligning human evaluation with the judgement of LLMs, we can iteratively specify the thresholds. I.e. we include humans in the loop serving as representatives of actual users to check and validate actual solutions and computed quality scores. Step-by-step we re-evaluate until the solutions are aligned with the preferences of the human evaluators. In particular, once humans judge solutions to be "good enough", the corresponding aligned $q_k$ and $v_k$ values serve as thresholds $\hat{q_k}$ and $\hat{v_k}$ for future evaluations.

The steps just explained can also be "formalized". In principle, when building applications which are supposed to be "good enough" one can stick to Algorithm~\ref{alg:determine-thresholds} including LLM-based and human evaluation components. It is shown at the conceptual level using pseudo-code as a guiding principle. We display relevant steps assuming that prompt engineering is employed for improving results. Analogous algorithms hold in case one uses finetuning or a mix of both\footnote{From a practical perspective, the algorithm may be extended with a termination criterion such as: “If, after $N$
iterations of prompt engineering or fine-tuning, no measurable improvement is observed for a given quality dimension, the current solution may be deemed ‘good enough’, or it may indicate the need to reconsider the application’s overall architecture.”}.
\begin{algorithm}[h!]
\footnotesize
\caption{Determine Thresholds}
\label{alg:determine-thresholds}
\begin{algorithmic}[1]
\vspace{1em}
\State \textbf{Input:}
\State \quad LLM\_System: The LLM model and its associated prompting strategy
\State \quad Problem\_Set: Representative set of problems
\State \quad Quality\_Dimensions: Trust index components (e.g., accuracy, tone) to be specified by domain experts
\Statex
\State \textbf{Output:}
\State \quad Thresholds: A set of pairs $(\hat{q}_k, \hat{v}_k)$ for each quality dimension $k$
\Statex
\Function{DetermineThresholds}{LLM\_System, Problem\_Set, Quality\_Dimensions}
    \State \textbf{Initialize:}
    \For{each $k \in$ Quality\_Dimensions}
        \State Set initial $\hat{q}_k=0$ and $\hat{v}_k=\infty$. \Comment{Represents the loosest possible thresholds}
    \EndFor
    \State Thresholds = []
    \Statex

    \State \textbf{Iterative Refinement Loop:}
    \Loop
        \State \textbf{a. Generate Solutions:}
        \State Current\_LLM\_Solutions = GenerateSolutions(LLM\_System, Problem\_Set)
        \Statex
        \State \textbf{b. Compute Quality Metrics:}
        \For{each solution $S$ in Current\_LLM\_Solutions}
            \For{each $k \in$ Quality\_Dimensions}
                \State $q_k(S), v_k(S)$ = ComputeMetrics(S, $k$, LLM\_System)
            \EndFor
        \EndFor
        \Statex
        \State \textbf{c. Human-in-the-Loop Validation:}
        \State Human\_Aligned\_Solutions = []
        \For{each solution $S$ in Current\_LLM\_Solutions}
            \If{HumanEvaluator.Validates(S, \{$q_k(S)$, $v_k(S)$ for all $k$\})}
                \State Add $S$ to Human\_Aligned\_Solutions
            \EndIf
        \EndFor
        \Statex
        \State \textbf{d. Update Thresholds \& Decide Next Step:}
        \If{HumanEvaluator.IsSatisfiedWith(Human\_Aligned\_Solutions, Problem\_Set)} \Comment{convergence for all $k$}
            \State \textbf{Compute Final Thresholds from Aligned Solutions:}
            \For{each $k \in$ Quality\_Dimensions}
                \Comment{Use conservative values from the set of approved solutions}
                \State $\hat{q}_k = \min \{q_k(S') \mid S' \in \text{Human\_Aligned\_Solutions}\}$
                \State $\hat{v}_k = \max \{v_k(S') \mid S' \in \text{Human\_Aligned\_Solutions}\}$
                \State Add $(\hat{q}_k, \hat{v}_k)$ to Thresholds
            \EndFor
            \State \textbf{Return} Thresholds
        \Else
            \Comment{Use human feedback to improve the system for the next iteration}
            \State Feedback\_Instances = Current\_LLM\_Solutions 
            \State LLM\_System = UpdatePrompts(LLM\_System, Feedback\_Instances)
        \EndIf
    \EndLoop
\EndFunction
\vspace{1em}
\end{algorithmic}
\end{algorithm}

\section{Ambiguity and Subjectivity}
In order to shift the discussion to a more practical level, we will now introduce two new quality dimensions (trust index components) which on the one hand illustrate the theoretical ideas presented here and which on the other hand fill a gap in the current landscape of LLM-based evaluation metrics.

In the literature multiple metrics for assessing LLM quality such as Answer Relevancy \cite{ragas}, Conversational Completeness or G-Eval \cite{geval} were already introduced. These metrics are generic and not bound to a specific dataset making them applicable to a broad range of possible use cases. Furthermore, the work of \cite{holisticevaluationlanguagemodels} introduces a top-down taxonomy-guided approach for benchmarking LLMs on a multitude of tasks allowing for a standardized approach towards evaluating use cases. 

While useful for many practical problems these ideas were not yet framed within the broader context of trust index and problem spaces. Furthermore, while these metrics already cover various aspects of language generation and understanding, they often do not take into account aspects such as semantic robustness or emotional evaluation criteria. We therefore introduce two new evaluation metrics which we shall call normalized bi-semantic entropy and emotional valence to fill this gap. 

\subsection{Normalized Bi-Semantic Entropy}
Ambiguity is a core feature of $P_{\text{NL}}$ and, consequently, of LLMs. We can formalize the diversity of meaning in potential responses and the robustness with respect to problem reformulations using the concept of normalized bi-semantic entropy. As we will see, this idea can be constructed on the basis of Shannon entropy, allowing for an information-theoretic interpretation.

Before defining normalized bi-semantic entropy, we remind the reader of a metric already introduced in existing literature \cite{kuhn2023semantic,farquhar2024detecting}.

\begin{definition}[Semantic Entropy SE]
Given a prompt $\tilde{q}$, we first generate a set of $N$ distinct answers, $A=\{A_1,A_2,...,A_N\}$. We then define a set of $M$ semantic classes, $C=\{C_1,C_2,...,C_M\}$, which represent possible categories that the answers could be assigned to. These semantic classes can either be defined by humans or be generated dynamically by invoking an LLM\footnote{Many applications building on the idea of multi-agent architectures rely on LLM-based orchestrators. Their task is to classify questions according to predefined categories motivated by actual use cases. For these orchestrators, semantic consistency is highly desirable.}. Using a semantic classifier, e.g. in terms of text embeddings or by invoking yet another LLM for classification, we assign each answer $A_i$ to one class $C_j$. Let $n_j$ be the count of answers assigned to class $C_j$. The probability of any given response falling into semantic class $C_j$ is $p_j = n_j/N$.
The semantic entropy for prompt $\tilde{q}$ is the Shannon entropy of this probability distribution:
$$\text{SE}(\tilde{q})=-\sum_{j=1}^{M}p_j\log_2 p_j$$
\begin{itemize}
    \item \textbf{Low SE}: Indicates semantic convergence. The LLM is "confident" about a narrow range of meanings (e.g., factual queries). The minimum is reached if $p_i = 1$ for $i \in {1, 2,..., M}$ and $p_j = 0$ for $p_j \neq p_i$.
    \item \textbf{High SE}: Indicates semantic divergence, exploring a wide variety of ideas. This is typical for ambiguous, open-ended, or creative prompts. Note that the maximum entropy is achieved for $p_i = p_j$ for all $i, j \in {1, 2,..., M}$, i.e. an answer falls into each semantic class with equal probability.
\end{itemize}
\textbf{Contrast with a Formal Program}: A deterministic formal program produces a single output, resulting in $\text{SE}=0$, the mathematical signature of zero ambiguity.
\end{definition}

While the above metric already captures semantic nuances of LLM-generated answers, the practical reality of human-LLM interaction is that users also express the same intent using different wordings. A robust LLM should be insensitive to superficial re-phrasings of the same underlying question. To measure this, we extend semantic entropy to bi-semantic entropy, which considers variations in the prompt itself.

\begin{definition}[Normalized Bi-Semantic Entropy SE$_{\text{Bi}}$]
We start by constructing the set of $K$ prompts, $\tilde{Q}=\{\tilde{q}_1,\tilde{q}_2,...,\tilde{q}_K\}$, that are semantically equivalent but phrased differently.
\begin{itemize}
    \item For each prompt $\tilde{q}_k \in \tilde{Q}$, generate a set of $N$ answers, $A_k=\{A_{k,1},A_{k,2},...,A_{k,N}\}$.
    \item Combine all answers into a total collection, $A_{\text{total}}=\bigcup_{k=1}^{K}A_k$. The total number of answers is $N_{\text{total}}=K \times N$.
    \item As before, define a set of $M$ emergent semantic classes, $C$, and classify each of the $N_{\text{total}}$ answers into one of the $M$ classes.
    \item Let $n_j$ be the total count of answers assigned to class $C_j$. The probability of any given response (to any of the prompt variants) falling into semantic class $C_j$ is $p_j = n_j/N_{\text{total}}$.
\end{itemize}
The bi-semantic entropy for the prompt set $\tilde{Q}$ is the Shannon entropy of this aggregate distribution:
$$\text{SE}_{\text{Bi}}(\tilde{Q})=-\sum_{j=1}^{M}p_j\log_2 p_j$$ In order to assess it in the unified language of solution quality introduced above we finally define its normalized version $$\text{NSE}_{\text{Bi}}(\tilde{Q}) = \frac{-\sum_{j=1}^{M} p_j \log_2 p_j}{\log_2 M}$$.
\begin{itemize}
    \item \textbf{Low NSE$_{\text{Bi}}$}: Indicates high robustness. The LLM's responses are semantically consistent regardless of how the question is phrased. This is desirable for factual queries and instruction-following. The minimum is reached if $p_i = 1$ for $i \in {1, 2,..., M}$ and $p_j = 0$ for $p_j \neq p_i$.
    \item \textbf{High NSE$_{\text{Bi}}$}: Indicates brittleness or sensitivity to prompt phrasing. The LLM gives semantically different answers to questions that have the same meaning. This can reveal how different phrasings trigger different biases or knowledge domains within the model. A powerful diagnostic use case is comparing $\text{SE}(\tilde{q}_k)$ for a single prompt with $\text{SE}_{\text{Bi}}(\tilde{Q})$. One might find that the model is highly consistent for one phrasing (low SE) but inconsistent across all phrasings (high SE$_{\text{Bi}}$). Note that the maximum entropy is achieved for $p_i = p_j$ for all $i, j \in {1, 2,..., M}$, i.e. an answer falls into each semantic class with equal probability.
\end{itemize}
\end{definition}
Connecting normalized bi-semantic entropy to our discussion of quality we can see that it nicely matches our requirements. We can compute it in multiple runs and assign mean value and variance. Depending on the use case we might prefer a high mean normalized bi-semantic entropy as well as a high variance (e.g. for open-ended and creativ tasks) or a low mean normalized bi-semantic entropy with low variance (e.g. factual tasks or classification). For the former the corresponding quality function could be chosen as $\text{NSE}_{\text{Bi}}(\tilde{Q})$ while for the latter it is $1 - \text{NSE}_{\text{Bi}}(\tilde{Q})$.

\subsection{Emotional Valence}
While the above quality metric operates at the semantic level a major piece that is missing in the current landscape of LLM evaluation metrics is a measure of emotional attachment with a given solution. For this purpose we introduce a new metric.

\begin{definition}[Emotional Valence $V_E$]
Let $A$ be an answer generated for a prompt $\tilde{q}$. For any individual human (or properly prompted LLM persona) $h$, their subjective interpretation of answer $A$ elicits an emotional response. We can model this with an emotional valence function:
$$V_E(A,h) \rightarrow [0, 1]$$
This function maps the answer to a scalar value, where values close to $1$ represent desirable emotional states (e.g. inspired, safe) and values close to $0$ represent undesirable states (e.g. offended, scared). A desirable strategy for an aligned AI system is to generate an answer $A$ that maximizes the expected emotional valence across a population $H$:
$$E[V_E(A)] = \frac{1}{|H|}\sum_{h \in H}V_E(A,h)$$
\end{definition}

\begin{definition}[Emotional Variance $\text{Var}(V_E)$]
The variance of the emotional valence measures how polarizing an answer is:
$$\text{Var}(V_E(A)) = E[(V_E(A)-E[V_E(A)])^2]$$
A high variance indicates a contentious answer that produces strong but divergent reactions. A low variance indicates a consensual response.
\end{definition}
In practice, depending on our needs, we might define several emotional valence scores. For example, we might want to optimize "excitement" as well as "amusement" when we design an LLM application which is supposed to generate thrilling but at the same time funny blog articles. 
\section{A Toy Model Application}
\label{sec:toy model}
In order to illustrate the conceptual ideas presented in this paper, we developed a simple toy model application\footnote{\href{https://github.com/RugTehseen/problem-solving-with-llms}{https://github.com/RugTehseen/problem-solving-with-llms}}. It is neither supposed to be production-ready software nor is it meant to truly simulate real-world use cases. Rather, it should be viewed as a tool for illustrating the concepts presented in this paper in an interactive, yet simplified environment. It can be viewed as a precursor and a starting-point guiding the construction of quality evaluation metrics in productive applications.

Concretely, we provide functionality for computing and adjusting emotional valence and normalized bi-semantic entropy in a simulated environment. In what follows we will briefly describe our implementation including an explanation how the application can be used to explore both concepts interactively. 

\begin{figure}[h!]
    \centering
    \includegraphics[width=0.89\textwidth]{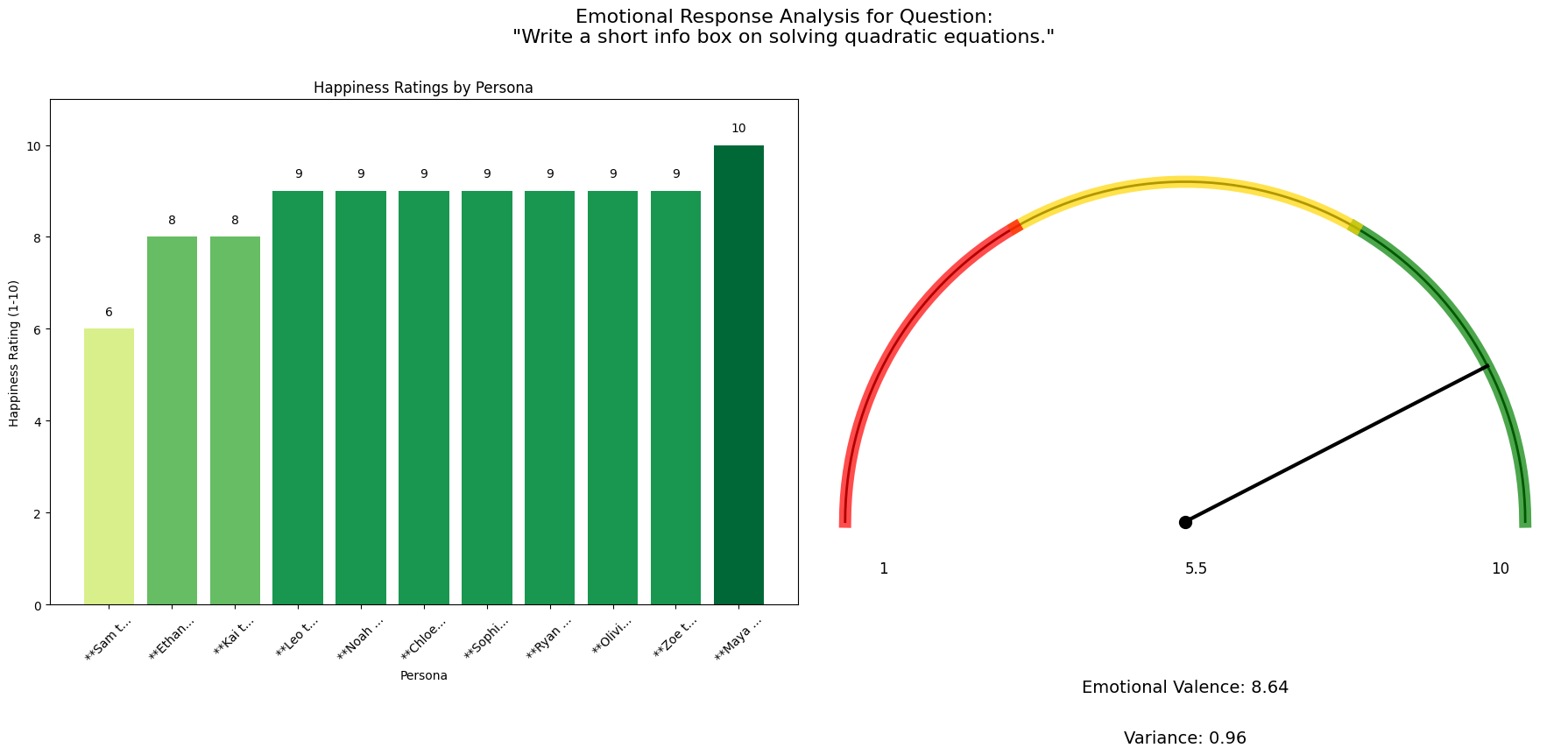} 
    \caption{Emotional valence and variance based on a "happiness" rating by LLM-proxies simulating a group of users of an app providing learning material for 9th graders. The image shows results based on an optimized system prompt. Further results can be found in the accompanying \href{https://github.com/RugTehseen/problem-solving-with-llms}{repository}.}
    \label{fig:emotional_valence}
\end{figure}

The application uses streamlit as frontend framework and ollama for integrating LLMs. During implementation we worked collaboratively with Claude-3.7 Sonnet as coding agent. Within the application, we are working with gemma3:4b albeit in principle one can use any LLM. We provide two python scripts, one encapsulating all logic for the emotional valence module, the other one containing all necessary code for evaluating normalized bi-semantic entropy. For evaluating the former, we include the following functionality
\begin{itemize}
    \item Customization of a use case that shall be explored, e.g. "Create an app for generating movie reviews"
    \item Automatic LLM-based generation of personas which serve as proxies of a real user base based on the use case definition
    \item Customization of a system prompt which is used when answering questions/addressing a problem
    \item Specification of a question or problem statement
    \item Generation of a solution by invoking the LLM
    \item Evaluation of the generated solution by all personas based on a "happiness" score ranging from 1 to 10
    \item Computation and visualization of emotional valence and variance on the basis of the evaluation results 
\end{itemize}
In order to analyze normalized bi-semantic entropy we provide
\begin{itemize}
    \item Specification of a question or problem statement
    \item Specification of categories that a solution could belong to
    \item Automated LLM-based semantic reframing of the original problem statement
    \item Multiple generations of an answer by LLM invocation for every semantic reformulation
    \item Categorization of each answer using an LLM-based classifier 
    \item Computation and visualisation of normalized bi-semantic entropy
\end{itemize}
Based on this functionality, we evaluated multiple examples from different domains. In the accompanying \href{https://github.com/RugTehseen/problem-solving-with-llms}{repository} we provide concrete results for two simple use cases, one from education and the other in the context of creative writing. The results include generated personas, emotional valence scores for different system prompts as well as evaluations of bi-semantic entropy for the system prompt that gave the best emotional valence scores. An illustrative emotional valence result in the context of education is shown in Figure~\ref{fig:emotional_valence}.

We recommend the user to explore other use cases and models and adjust the metrics and evaluations according to individual needs. 

\section{Conclusion and Future Work}
The emergence of LLMs necessitates a rethinking of what we mean by "computation" and "solution". By moving beyond a purely logic-based view, we can appreciate the new frontier of problem-solving that AI is now entering.

This paper has proposed a formal framework to navigate this new terrain. By delineating problem spaces $P_{\text{Formal}}$, $P_{\text{NL}}$ and $P_{\text{LLM}}$ and introducing a unified measure for solution quality, we can move beyond simple accuracy metrics to quantify the more nuanced aspects of linguistic solutions. 
The work thereby established concrete computational and statistical means for evaluating and finding "good enough" solutions guiding the path towards practical implementation within a unified framing.
In addition, we introduced two new quality metrics explicitly designed to measure semantic variability as well as emotional response inherent to an LLM's solution and illustrated these ideas using simple examples.

This is a starting point. Among others, future work should focus on:
\begin{itemize}
    \item Developing scalable methodologies for measuring and operationalizing the trust index \( Q \), potentially through advanced LLM-as-a-Judge architectures.
    \item Implementing the proposed algorithm for identifying “good enough” solutions in real-world application domains.
    \item Refining and formalizing the boundaries between problem spaces through extended theoretical analysis.
    \item Investigating the cognitive and collaborative dynamics of human–LLM teams in solving complex problems within \( P_{\text{NL}} \).
    \item Empirically evaluating the proposed quality metrics in practical deployments and developing new, task-specific evaluation criteria.
\end{itemize}
In conclusion, by explicitly embracing natural language as a medium for problem-solving---and by formalizing concepts such as problem spaces, trust index, ambiguity, and subjectivity---we can lay the foundation for a more mature science of LLM-based systems.

Such a science is better equipped to move beyond the constraints of purely logical computation, enabling digital systems to engage with problems that are expressible only through natural language. This paradigm shift opens up new frontiers in the digitalization of language-intensive domains, made possible by the evolving capabilities of large language models.

\section*{Acknowledgements}

The authors would like to thank Henry Heitmann and Jürgen Lind for valuable discussions and helpful feedback during the development of this work.
\bibliographystyle{unsrt}
\bibliography{references}
\end{document}